\RequirePackage{luatex85}

\documentclass[twoside,leqno,twocolumn]{article}
\usepackage{ltexpprt}

\usepackage{acronym}
\usepackage{algorithm} 
\usepackage{algorithmicx} 
\usepackage[noend]{algpseudocode} 
\usepackage{amsmath,amssymb}
\usepackage{booktabs}
\usepackage[skip=8pt]{caption} 
\usepackage{enumitem}
\usepackage{graphicx}
\usepackage{mathrsfs}
\usepackage{placeins}
\usepackage{siunitx}
\usepackage{subcaption} 
\usepackage{xspace}

\MakeRobust{\Call}

\usepackage[dvipsnames]{xcolor}
\usepackage{tikz}
\usetikzlibrary{}
\usepackage{tikzscale}
\usepackage{pgfplots}
\pgfplotsset{compat=newest}
\pgfplotsset{every axis legend/.append style={%
cells={anchor=west}}
}
\usepgfplotslibrary{polar}
\usetikzlibrary{shapes,arrows,fit}
\tikzset{>=stealth'}
\usepgfplotslibrary{groupplots}

\usetikzlibrary{patterns}
\usetikzlibrary{positioning}
\usetikzlibrary{arrows.meta, calc, shapes}

\tikzset{%
  >={Latex[width=2mm,length=2mm]},
            base/.style = {rectangle, rounded corners, draw=black,
                           minimum width=1cm, minimum height=1cm,font={\Large,\bf},inner sep=0.3cm,
                           fill=teal!30,
                           text centered},
            termstyle/.style = {base, very thick}, 
            nontermstyle/.style = {base, draw=none, font={\Large}},
            tfstyle/.style = {font={\Large}},
            explainstyle/.style = {font={\Large},text=blue, align=left},
}
\pgfdeclarelayer{background}
\pgfdeclarelayer{foreground}
\pgfsetlayers{background,main,foreground}

\def\nodedist{2cm}

\usepackage{aircraftshapes}
\usepackage{hyperref}
\usepackage{cleveref} 

\setlist{noitemsep,topsep=0.5ex}
\newcommand{\argmin}{\operatornamewithlimits{arg\,min}}

\algnewcommand\And{\textbf{and}\xspace}
\algnewcommand\Or{\textbf{or}\xspace}
\algnewcommand{\LineComment}[1]{\State \(\triangleright\) #1}

\newacro{nmac}[NMAC]{near mid-air collision}
\acrodefindefinite{nmac}{an}{a}
\newacro{mcts}[MCTS]{Monte Carlo tree search}
\newacro{mctsdpw}[MCTS-DPW]{Monte Carlo Tree Search with Double Progressive Widening}
\newacro{uct}[UCT]{Upper Confidence Tree}
\newacro{acasx}[ACAS X]{Airborne Collision Avoidance System}
\newacro{tcas}[TCAS]{Traffic Alert and Collision Avoidance System}
\newacro{mdp}[MDP]{Markov Decision Process}
\newacro{nasa}[NASA]{National Aeronautics and Space Administration}
\newacro{faa}[FAA]{Federal Aviation Administration}
\newacro{icao}[ICAO]{International Civil Aviation Organization}
\newacro{ra}[RA]{resolution advisory}
\acrodefindefinite{ra}{an}{a}
\acrodefplural{ra}[RAs]{resolution advisories}
\newacro{ta}[TA]{traffic alert}
\newacro{llcem}[LLCEM]{Lincoln Laboratory Correlated Aircraft Encounter Model}
\newacro{fteg}[FTEG]{Fast-Time Encounter Generator}
\newacro{mitll}[MIT-LL]{MIT Lincoln Laboratory}
\newacro{apl}[JHUAPL]{Johns Hopkins University Applied Physics Laboratory}
\newacro{saso}[SASO]{Safe and Autonomous Systems Operations}
\newacro{aosp}[AOSP]{Airspace Operations and Safety Program}

\newacro{arm}[ARM]{associative rule mining}
\newacro{cart}[CART]{classification and regression tree}
\newacro{cpa}[CPA]{closest point of approach}
\newacro{ge}[GE]{grammatical evolution}
\newacro{ga}[GA]{genetic algorithm}
\newacro{gp}[GP]{genetic programming}
\newacro{cfg}[CFG]{context-free grammar}
\newacro{bnf}[BNF]{Backus-Naur Form}
\newacro{gbdt}[GBDT]{grammar-based decision tree}
\newacro{gbes}[GBES]{grammar-based expression search}
\newacro{ilp}[ILP]{inductive logic programming}
\newacro{ltl}[LTL]{linear temporal logic}
\newacro{mtl}[MTL]{metric temporal logic}
\newacro{sax}[SAX]{symbolic aggregate approximation}

\newacro{uci}[UCI]{UC Irvine}

\begin{document}

\title{\Large Interpretable Categorization of Heterogeneous Time Series Data}
\author{Ritchie Lee\thanks{Carnegie Mellon University Silicon Valley.} \and
Mykel J. Kochenderfer\thanks{Stanford University.} \and
Ole J. Mengshoel\thanks{Carnegie Mellon University Silicon Valley.} \and
Joshua Silbermann\thanks{Johns Hopkins University Applied Physics Laboratory.}}
\date{}

\maketitle


\fancyfoot[R]{\footnotesize{\textbf{Copyright \textcopyright\ 2018 by U.S. Government\\
Unauthorized reproduction of this article is prohibited}}}





\begin{abstract} \small\baselineskip=9pt Understanding heterogeneous multivariate time series data is important in many applications ranging from smart homes to aviation.  Learning models of heterogeneous multivariate time series that are also human-interpretable is challenging and not adequately addressed by the existing literature.  We propose grammar-based decision trees (GBDTs) and an algorithm for learning them.  GBDTs extend decision trees with a grammar framework.  Logical expressions derived from a context-free grammar are used for branching in place of simple thresholds on attributes.  The added expressivity enables support for a wide range of data types while retaining the interpretability of decision trees.  In particular, when a grammar based on temporal logic is used, we show that GBDTs can be used for the interpretable classification of high-dimensional and heterogeneous time series data.  Furthermore, we show how GBDTs can also be used for categorization, which is a combination of clustering and generating interpretable explanations for each cluster.  We apply GBDTs to analyze the classic Australian Sign Language dataset as well as data on near mid-air collisions (NMACs).  The NMAC data comes from aircraft simulations used in the development of the next-generation Airborne Collision Avoidance System (ACAS X).  \end{abstract}

\section{Introduction.} 
Heterogeneous multivariate time series data arises in many applications including driverless cars, smart homes, robotic servants, and aviation.  Understanding these datasets is important for designing better systems, validating safety, and analyzing failures.  However, knowledge discovery in heterogeneous multivariate time series datasets is very challenging because it typically requires two major data mining problems to be addressed simultaneously.  The first problem is how to handle multivariate \emph{heterogeneous} time series data, where the variables are a mix of numeric, Boolean, and categorical types.  The second problem is the need for \emph{interpretability}.  That is, humans must be able to understand and reason about the information captured by the model.  While these problems have been explored separately in the literature, we are not aware of any methods that address both interpretability and heterogeneous multivariate time series datasets together.

Rule-based methods such as decision trees \cite{Breiman1984} and decision lists \cite{Rivest1987} are very intuitive because they use symbolic rules for decision boundaries.  However, they do not support time series data.  Motif discovery methods, such as shapelet \cite{Ye2011} and subsequence \cite{Senin2013} discovery, find recurring patterns in sequential data.  These methods are less interpretable than rule-based methods as they report prototypes of patterns rather than state relationships.  These methods have also not been applied to heterogeneous multivariate time series data.  On the other hand, models that support heterogeneous multivariate time series, such as recurrent neural networks \cite{Gers2000}, are not interpretable. 

To simultaneously address the problems of interpretability and heterogeneous time series data, we increase the expressiveness of decision trees by allowing decision rules to be any logical expression.  Traditional decision trees partition the input space using simple thresholds on attributes, such as $(x_1 < 2)$ \cite{Breiman1984}.  However, these partitioning rules have limited expressiveness and cannot be used to express more complex logical relationships, such as those between heterogeneous attributes or across time.  Allowing more expressive rules in the decision tree increases its modeling capability while retaining interpretability.  In the past, first order logic rules have been used in decision trees \cite{Blockeel1998}.  We generalize this idea and allow any logical language that can be specified using a context-free grammar.  In particular, we show that decision trees combined with temporal logic can be very effective as an interpretable model for heterogeneous multivariate time series data.  Temporal logic, combined with other methods, have been applied to heterogeneous time series data in aviation \cite{Schumann2013}. 


The key contributions of this paper are:
\begin{itemize}
    \item a \ac{gbdt} framework that combines decision trees with a grammar;
    \item a training algorithm for \ac{gbdt} that leverages existing \ac{gbes} algorithms, such as \ac{gp} \cite{Koza1992}  and \ac{ge} \cite{ONeil2003};
    \item an alternative view of the \ac{gbdt} model as categorization, which is the combination of clustering data into similar groups and providing explanations for them; and
    \item experiments with \acp{gbdt} and temporal logic on two datasets, showing that \acp{gbdt} learn reasonable explanations while producing competitive classification performance compared to baselines.
\end{itemize}

%
\section{Related Work.}
\label{sec:related}

%
\subsection{Interpretable Models.} A variety of interpretable models for static data have been proposed in the literature.  Regression models \cite{Schielzeth2010}, generalized additive models \cite{Lou2012}, and Bayesian case models \cite{Kim2014} have been recently proposed as models with interpretability.  These models aid interpretability by stating decision boundaries in terms of basis functions or representative instances (prototypes).  Bayesian networks have also been used for prediction and data understanding \cite{Basak2016}.  Our approach aims to achieve better interpretability by representing decision boundaries symbolically using (temporal) logic rules rather than in terms of basis functions, examples, or statistical weights.

Rule-based models, such as decision trees \cite{Breiman1984}, decision lists \cite{Rivest1987}, and decision sets \cite{Lakkaraju2016}, are easy to understand because their decision boundaries are stated in terms of input attributes and simple logical operators.  
These methods provide good interpretability, but do not capture the temporal aspect of time series data. 

%
%
%
\subsection{Time Series Models.} Time series analysis has focused on dynamic time warping, hidden Markov models, dictionary-based approaches, and recurrent neural networks \cite{Kadous2002}\cite{Bagnall2017}\cite{Gers2000}.  Shapelets \cite{Ye2011} and subsequence search \cite{Senin2013} have been proposed for univariate time series classification.  These approaches search for simple patterns that are most correlated with the class label.  Interpretability comes from identifying a prototype of a recurring subsequence pattern.  Implication rules \cite{Shokoohi2015} and simple logical combinations of shapelets \cite{Mueen2011} have been proposed to extend the shapelets approach.  Learning prototypes or statistical weights are not as interpretable as symbolic rules.  

\subsection{Grammars and Decision Trees.} The combination of decision trees and grammars has been proposed in the past \cite{Wong2006}\cite{Motsinger2010}.  These prior works use a grammar to optimize traditional decision trees as a whole where the splits are simple thresholds over individual attributes.  The resulting tree is a traditional decision tree.  In contrast, \ac{gbdt} uses a grammar to optimize node splits and the resulting decision tree contains a (temporal) logic rule at each node split.  This distinction is important because the added expressivity of the logical expressions is what allows support for multivariate heterogeneous time series datasets.  

\section{Preliminaries.}
\label{sec:prelims}

\subsection{Notation.}

A multi-dimensional time series dataset $D$ consists of $m$ record-label pairs $((r_1,l_1),(r_2,l_2),...(r_m,l_m))$, where a record $r$ is a two-dimensional matrix of $n$ attributes by $T$ time steps and a label $l$ is its discrete class label.  A trace $\vec{x}_i$ is the $i$'th row of a record and represents the time series of that attribute.  Logical and comparison operators are given broadcast semantics where appropriate.  For example, the comparison operator in $\vec{x}_i < c$ compares each element of $\vec{x}_i$ to $c$ and returns a vector of the same size as $\vec{x}_i$.  Similarly, the logical operator in $\vec{x}_i \wedge \vec{x}_j$ operates elementwise.  The temporal operators $\text{F}$ and $\text{G}$ are \textit{eventually} and \textit{globally}, respectively. \textit{Eventually} returns true if any value in the argument vector is true.  \textit{Globally} returns true if all values in the argument vector are true. 

\subsection{Context-Free Grammar.}

A \acf{cfg} defines a set of rules that govern how to form expressions in a formal language, such as \ac{ltl} \cite{Gabbay1980}. The grammar defines the syntax of the language and provides a means to generate valid expressions.  A \ac{cfg} $G$ is defined by a 4-tuple $(\mathscr{N}, \mathscr{T}, \mathscr{P}, \mathscr{S})$, where $\mathscr{N}$ is a set of \textit{non-terminals}; $\mathscr{T}$ is a set of \textit{terminals}; $\mathscr{P}$ is a set of \textit{production rules}, which are rules governing the substitution of non-terminals;
and $\mathscr{S}$ is the special \textit{start} symbol that begins a derivation.  
The derivation is commonly represented as a tree structure called a derivation tree.  
We assume that the semantics of the symbols in the grammar are defined.

%
\subsection{Grammar-Based Expression Search.}
\label{sec:exprsearch}

Grammar-based expression search (GBES)\acused{gbes} is the problem of finding an expression $e^*$ from a grammar that minimizes a given fitness function $f(e)$, i.e., $e^*=\argmin_e{f(e)}$ \cite{McKay2010}.  The formulation is extremely general due to the flexibility and expressiveness of grammars and the arbitrary choice of fitness function.  
We describe three existing \ac{gbes} algorithms as follows.  

\textbf{Monte Carlo.} Monte Carlo generates expressions by repeatedly selecting non-terminals in the partial expression and applying a production rule chosen uniformly at random amongst possible rules.  When no non-terminals remain, the fitness of the generated expression is evaluated and the expression with the best fitness is reported.  A maximum depth is typically used to ensure that the process terminates.  

\textbf{Grammatical Evolution.} Grammatical evolution (GE)\acused{ge} \cite{ONeil2003} is an evolutionary algorithm that is based on a sequential representation of the derivation tree.  Specifically, \ac{ge} defines a transformation from a variable-length integer array to a sequence of rule selections in a \ac{cfg}. Then it uses a standard \ac{ga} to search over integer arrays.  We use one-point crossover and uniform mutation \cite{Poli2008}.  


\textbf{Genetic Programming.} Genetic programming (GP)\acused{GP} is an evolutionary algorithm for optimizing trees \cite{Whigham1995}.  Genetic operators are defined specifically for trees and thus do not require any transformations of the derivation tree.  Our implementation uses a crossover operator that exchanges compatible subtrees between two individuals and a mutation operator that replaces entire subtrees with randomly-generated ones. We use tournament selection for selecting individuals \cite{Koza1992}.

\acresetall
\section{Grammar-Based Decision Trees.}
\label{sec:gbdt}

Grammar-based decision trees (GBDTs)\acused{gbdt} extend decision trees with a grammar framework to allow general logical expressions to be used as the branching rules in a decision tree.  The domain of the logical expressions is constrained using a \ac{cfg}.  In this paper, we consider grammars based on temporal logic for the classification of heterogeneous multivariate time series data.  The grammar can be easily adapted to the characteristics of the data or the application.  

\subsection{Model.}
\ac{gbdt} is a binary decision tree where each non-leaf node contains a logical expression.  Each branch from a non-leaf node is associated with a possible outcome of the logical expression at that node.  Leaf nodes are associated with a particular class label for prediction.  
We use the following notation: A \ac{gbdt} tree $\mathcal{T}$ is either a leaf with class $l$, which we denote by $\mathcal{T}=\textsc{Leaf}(l)$; or it is an internal node with expression $e \in G$, true branch $\mathcal{T}^+$, and false branch $\mathcal{T}^-$, which we denote by $\mathcal{T}=\textsc{INode}(e, \mathcal{T}^+, \mathcal{T}^-)$.  

\textbf{Prediction.} As in a traditional decision tree, prediction is performed in a top-down fashion starting at the root.  At each node, the logical expression is evaluated on the record and the result dictates which branch is followed.  The process continues recursively until a leaf node is reached where the class label at the leaf node is used for prediction. 

\textbf{Categorization.} While decision trees are traditionally used for classification, they can also be used for categorization, which is the combined task of clustering and explaining data.  A categorization of the data can be extracted from a \ac{gbdt} by considering each leaf node of the tree to be a separate category.  The description of the category is then the conjunction of all branch expressions between the root and the leaf node of interest.  The members of the cluster are the records where the cluster's description holds true.  Since partitions are mutually exclusive, the clusters do not overlap.   

\begin{figure}[ht!]
    \centering
    \begin{subfigure}[b]{0.5\columnwidth}
        \centering
        \includegraphics[width=1.0\columnwidth]{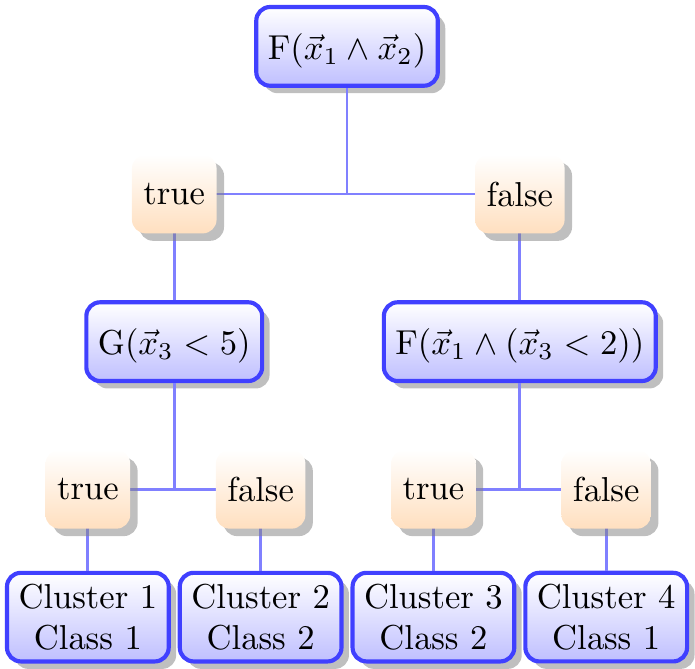}
        \caption{Tree}
        \label{fig:gbdtexample1}
    \end{subfigure}%
    \begin{subfigure}[b]{0.4\columnwidth}
        \scriptsize
        \centering
            \begin{align*}
                \textsc{Start} &::= \textsc{B}\\
                \textsc{B} &::= \textsc{Ev} \mid \textsc{Gl}\\
                \textsc{Ev} &::= \text{F}(\textsc{VB})\\
                \textsc{Gl} &::= \text{G}(\textsc{VB})\\
                \textsc{VB} &::= \vec{x}_1 \mid \vec{x}_2 \mid \textsc{And} \mid \textsc{Lt}\\
                \textsc{And} &::= \textsc{VB} \wedge \textsc{VB}\\
                \textsc{Lt} &::=  \textsc{VR} < C\\
                \textsc{VR} &::= \vec{x}_3 \mid \vec{x}_4\\
                \textsc{C} &::= 1 \mid 2 \mid 3 \mid 4 \mid 5
            \end{align*}
        \caption{Grammar}
        \label{fig:gbdtexample2}
    \end{subfigure}
    \caption{Grammar-based decision tree example.}
    \label{fig:gbdtexample}
\end{figure}

\textbf{Example.} \Cref{fig:gbdtexample} shows an example of a \ac{gbdt} with an accompanying \ac{cfg} expressed in \acf{bnf}.  The \ac{cfg} describes a simple temporal logic.  It assumes that the data has four attributes $\vec{x}_1,\vec{x}_2,\vec{x}_3,\vec{x}_4$, where the attributes $\vec{x}_1$ and $\vec{x}_2$ are Boolean vectors and $\vec{x}_3$ and $\vec{x}_4$ are vectors of real numbers.  The grammar contains two types of rules.  \textit{Expression rules}, such as $\textsc{Ev} ::= \text{F}(\textsc{VB})$, contain partially-formed expressions that contain terminal and non-terminal symbols.  Non-terminal symbols are substituted further using the appropriate production rules.  \textit{Or Rules}, such as $B ::= \textsc{Ev} \mid \textsc{Gl}$, contain the symbol $\mid$, which delineates the different possible substitutions.   

Each non-leaf node of the \ac{gbdt} contains a logical expression derived from the grammar.  The expression dictates which branch should be followed.  Leaf nodes show the predicted class label.  In \Cref{fig:gbdtexample}, a data record where $\text{F}(\vec{x}_1 \wedge \vec{x}_2)$ is true and $\text{G}(\vec{x}_3 < 5)$ is false would be predicted to have class label 2.  We also label each leaf node with a unique cluster number.  For example, the righmost leaf node in \Cref{fig:gbdtexample1} is labeled as cluster 4.  The overall expression that describes cluster 4 is $\neg \text{F}(\vec{x}_1 \wedge \vec{x}_2) \wedge \neg \text{F}(\vec{x}_1 \wedge (\vec{x}_3 < 2))$. 

\subsection{Grammars for Heterogeneous Time Series.}

%
In the \ac{gbdt} framework, logic expressions are evaluated on a data record and produce a Boolean output. The symbols of the expression can refer to fields of the record, constants, or functions.  We adopt a subset of \acf{ltl}, a formal logic with temporal operators often used in the analysis of time series data \cite{Gabbay1980}.  Temporal logic provides an elegant framework for defining expressions that support heterogeneous multivariate time series data.  For example, the logic rule $\text{F}(\vec{x}_1 \wedge (\vec{x}_3 < 2))$ in \Cref{fig:gbdtexample} contains a Boolean vector $\vec{x}_1$ and a vector of real values $\vec{x}_3$.  The rule integrates these two different data types elegantly by first evaluating $(\vec{x}_3 < 2)$ using an elementwise comparison to produce a Boolean vector.   Then, the result is combined elementwise via conjunction with $\vec{x}_1$ to produce another Boolean vector.  Finally, the temporal operator $\text{F}$ collapses the Boolean vector to a single Boolean output.  The constants in the grammar can be preselected based on discretizing over the range of each real-valued attribute in the dataset.  Alternatively, some \ac{gbes} algorithms, such as Monte Carlo and genetic programming, also support drawing random values during search \cite{Koza1992}.


For simplicity, the example in \Cref{fig:gbdtexample} included only a small number of operators.  The grammar can be readily extended to include other operators including disjunct $\vee$, greater than $>$, equals $=$, where equality is important for categorical attributes, and even arbitrary user-defined functions.  Depending on the application, a generic temporal logic grammar can be used (as we have done with the Auslan dataset) or the grammar can be tailored to the application (as we have done with the \acs{acasx} dataset).  This paper considers very simple temporal operators that operate over the entire time series.  More sophisticated temporal logics, such as \ac{mtl} \cite{Koymans1990}, can be used with \ac{gbdt} to discover more intricate patterns in the data. 

\subsection{Natural Language Descriptions.}

Logical expressions can sometimes be dense and hard to parse.  In many cases, we can improve interpretability by providing natural language descriptions of the expressions.  One method to automatically translate expressions into English sentences is to map expression rules and terminal symbols in the \ac{cfg} to corresponding sentence fragments and then use the structure of the expression's derivation tree to assemble the fragments.  \Cref{fig:naturallangexample} shows an example mapping that could be used with the grammar in \Cref{fig:gbdtexample}.  

\begin{figure}[!ht]
    \centering
    \footnotesize
        \begin{align*}
            \text{F}(\textsc{VB}) &:= \text{``at some point, $\textsc{VB}$"}\\
            \text{G}(\textsc{VB}) &:= \text{``for all time, $\textsc{VB}$"}\\
            \textsc{VB} \wedge \textsc{VB} &:= \text{``$\textsc{VB}$ and $\textsc{VB}$}"\\
            \textsc{VR} < C &:= \text{``$\textsc{VR}$ is less than $C$"}\\
            \vec{x}_1 &:= \text{``advisory is active"}\\
            \vec{x}_2 &:= \text{``pilot is responding"}\\
            \vec{x}_3 &:= \text{``vertical rate"}\\
            \vec{x}_4 &:= \text{``airspeed"}
        \end{align*}
    \caption{Natural language map example.}
    \label{fig:naturallangexample}
\end{figure}

Applying the mapping in \Cref{fig:naturallangexample}, the decision rules in \Cref{fig:gbdtexample} become the following natural language descriptions:  $\text{F}(\vec{x}_1 \wedge \vec{x}_2)$ is translated to ``at some point, [advisory is active] and [pilot is responding]"; 
$\text{G}(\vec{x}_3 < 5)$ is translated to ``for all time, [vertical rate] is less than 5"; and
$\text{F}(\vec{x}_1 \wedge (\vec{x}_3 < 2))$ is translated to ``at some point, [advisory is active] and [[vertical rate] is less than 2]".  We include square brackets to help the reader disambiguate nested sentence components.  

\section{Induction of GBDTs.}
\label{sec:gbdtinduction}

Induction of a \ac{gbdt} is performed top-down as in traditional decision tree learning.  However, there is a difference in our use of \ac{gbes} as a subroutine to find the best partition expression.  The induction algorithm begins with a single (root) node containing the entire dataset. \ac{gbes} is then used to search a \ac{cfg} for the partitioning expression that yields the best fitness.  The expression is evaluated on each record and the dataset is partitioned into two child nodes according to the results of the evaluation.  The process is applied recursively to each child until all data records at the node are either correctly classified or a maximum tree depth is reached.  The mode of the training labels is used for class label prediction at a leaf node. The \ac{gbdt} induction algorithm is shown in \Cref{alg:gbdt-ind}.

\begin{algorithm}
    \footnotesize
    \caption{\label{alg:gbdt-ind}Grammar-Based Decision Tree Induction}
    \begin{algorithmic}[1]
        \LineComment{Inputs: \ac{cfg} $G$, Fitness Function $f$, Dataset $D$, Depth $d$}
        \Function{GBDT}{$G, f, D, d$}
            \State $R \gets \Call{Split}{G, f, D, d}$
            \State \Return $\Call{Tree}{R}$ 
        \EndFunction
        \Function{Split}{$G, f, D, d$}
            \If{$\Call{IsHomogeneous}{\Call{Labels}{D}}$ \Or $d = 0$}
                \State \Return \Call{Leaf}{\Call{Mode}{\Call{Labels}{D}}} 
            \EndIf 
            \State $\hat{e}^* \gets \Call{GBES}{G, f, D}$
            \State $(D^+, D^-) \gets \Call{SplitData}{D, \hat{e}^*}$ 
            \State $child^+ \gets \Call{Split}{G, f, D^+, d - 1}$
            \State $child^- \gets \Call{Split}{G, f, D^-, d - 1}$
            \State \Return $\Call{INode}{\hat{e}^*, child^+, child^-}$
        \EndFunction
    \end{algorithmic}
\end{algorithm}

In \Cref{alg:gbdt-ind}, \textsc{GBDT} (line 3) is the main entry point to the induction algorithm.  It returns a \textsc{Tree} object containing the root node of the induced decision tree. \textsc{Split} (line 6) attempts to partition the data into two parts.  It first tests whether the terminal conditions are met and if so returns a \textsc{Leaf} object that predicts the mode of the labels.  The partitioning terminates if the maximum depth has been reached or if all class labels are the same, which is tested by the \textsc{IsHomogeneous} function (line 7). The \textsc{GBES} function (line 9) uses \ac{gbes} to search for the expression that minimizes the fitness function $f(e)$, i.e., $e^*=\argmin_e{f(e)}$.  \textsc{SplitData} (lines 12-13) evaluates the expression on each record and partitions the data into two parts according to whether the expression holds.  Then, \textsc{Split} (lines 12-13) is called recursively on each part. \textsc{Split} returns an \textsc{INode} object containing the decision expression and the children of the node.

\subsection{Fitness Function.}
\label{sec:fitness}

In \textsc{GBES} (line 9), we evaluate the desirability, or \textit{fitness}, of an expression according to two competing objectives.  On the one hand, we want expressions that partition the data so that the resulting leaf nodes have the same ground truth class labels.  Splits that induce high homogeneity tend to produce shallower trees and thus shorter global expressions at leaf nodes.  They also produce classifiers with better predictive accuracy when the maximum tree depth is limited.  To quantify homogeneity, we use the Gini impurity metric following the \ac{cart} framework \cite{Breiman1984}.  On the other hand, we want to encourage interpretability by minimizing the length and complexity of expressions.  Shorter and simpler expressions are generally easier to interpret.  We use the number of nodes in the derivation tree as a proxy for the complexity of an expression $e$.  The two objectives are combined linearly into a single (minimizing) fitness function given by 
\begin{equation*}
    f(e) = w_1 I_G + w_2 N_e
\end{equation*}
where $f$ is the fitness function, $w_1 \in \mathbb{R}$  and $w_2 \in \mathbb{R}$ are weights,  and $N_e$ is the number of nodes in the derivation tree of $e$.  The total Gini impurity, $I_G$, is the sum of the Gini impurity of each partition that results from splitting the data using expression $e$.  It is given by
\begin{equation*}
    I_G = \sum_{L \in \{L^+,L^-\}}\sum_{b \in B}f^b_L (1-f^b_L)
\end{equation*}
where $f^b_L$ is the fraction of labels in $L$ that are equal to $b$; $L^+$ are the labels of the records on which $e$ evaluates to true, i.e., $L^+=[l \mid \textsc{Evaluate}(e,r), (r,l) \in D]$; $L^-$ are the labels of the records on which $e$ evaluates to false, i.e., $L^-=[l \mid \neg\textsc{Evaluate}(e,r), (r,l) \in D]$; and $B = \{$True, False$\}$.  We use square brackets with set construction syntax to indicate that $L^+$ and $L^-$ are vectors and can have duplicate elements.

\subsection{Computational Complexity.}
The most computationally expensive part of \ac{gbdt} training is evaluating the fitness of an expression since it involves visiting each record in the dataset and then computing statistics.  \ac{gbes} also requires a large number of expression evaluations to optimize the decision expression at each decision node.  The deeper the tree, the more nodes need to be optimized.  However, as the tree gets deeper, the nodes operate on increasingly smaller fractions of the dataset.  In fact, while the number of nodes grows exponentially with tree depth, the number of records that must be evaluated at each level remains constant (the size of the dataset).  Overall, the computational complexity of \ac{gbdt} induction is $O(m \cdot N_{GBES} \cdot d)$, where $m$ is the number of records in the dataset, $N_{GBES}$ is the number of logical expressions evaluated in each \ac{gbes}, and $d$ is the depth of the decision tree.

\section{Experiments.} 
We evaluate \acp{gbdt} on the Australian Sign Language dataset and data from an aircraft encounter simulator.  Both experiments use the same metrics and baselines.

\textbf{Metrics.} For classification performance, we report the accuracy, precision, recall, and F1 scores.  We use `macro' averaging, which calculates the mean of the binary metrics, giving equal weight to each class. For interpretability, we report the average number of terminals in each rule of the decision tree.  For example, $\text{F}(\vec{x}_1 \wedge \vec{x}_2)$ has four terminals: $\text{F}$, $\vec{x}_1$, $\wedge$, and $\vec{x}_2$.  Rules containing fewer terminals are considered easier to interpret.  A \textit{trial} is a run of an experiment initialized with a unique random seed and train-test split.  We perform random trials using stratified randomly-sampled 70/30 train-test splits of the data and report the metric's mean over the trials on the testing set.

\textbf{Baselines.} We compare our algorithm against decision tree, random forest, and Gaussian naive Bayes classifiers from Scikit-Learn \cite{Scikitlearn}.  Random forest and Gaussian naive Bayes models have low interpretability.  We preprocess the data before applying the baseline algorithms as follows: z-score normalize each real-valued attribute; convert Boolean attributes to [-1.0,1.0]; zero-pad so that all records have the same length in the time dimension; and reshape the time series data (size $m \times n \times T$) into non-time series data (size $m \times nT$) by concatenating time slices.  \ac{gbdt} can directly handle heterogeneous multivariate time series data of varying length and thus does not require this preprocessing.  

\subsection{Australian Sign Language.}
\label{sec:auslan}
We analyze the classic Australian Sign Language (``Auslan'') dataset from the \ac{uci} repository \cite{UCI}\cite{Kadous2002}.  The data originates from participants wearing instrumented gloves while signing specific words in Australian sign language. The dataset is a multivariate time series of varying episode length containing 22 continuous attributes, 95 words with 27 examples each. We extract eight words (classes) from the dataset: \textit{hello, please, yes, no, right, wrong, same, different}.

\textbf{Grammar.} We use a generic temporal logic grammar that includes all attributes and the following operators: $\text{F}$, $\text{G}$, $\implies$, $\neg$, $\vee$, $\wedge$, $=$, $<$, $\leq$, $>$, $\geq$.  Since attributes may have different ranges, constant values used in comparison operators must be specialized to each attribute.  To address this issue, we consider the range of each attribute in the dataset and uniformly discretize it into 10 points.  Attributes are compared with their corresponding set of discretized values.  The data itself is not being discretized.  Discretization is only used to generate the threshold constants in the grammar. 

\subsubsection{Results.} We performed 20 random trials for each algorithm and report the mean results in Table \ref{tab:auslanmetrics}. \ac{gbdt}, decision tree, and random forest models were trained to a maximum depth $d=6$ and random forest used 32 trees.  \ac{gbdt}-GP performed the best overall both in terms of classification and interpretability metrics.  
Decision tree performed poorly on this dataset because key characteristics of the attributes may be shifted in time due to variations in the speed of signing.  Random forest can partially overcome this issue by using a larger number of trees.  The temporal logic rules in \ac{gbdt} can more effectively capture the underlying temporal properties of the data. 

\begin{table*}[!ht]
    \centering
    \footnotesize
    \caption{Algorithm performance metrics on Auslan dataset. Best value for each metric is in bold.}
        \begin{tabular}{@{}lcccccc}
            \toprule
            \textbf{} & \textbf{GBDT-MC} & \textbf{GBDT-GE} & \textbf{GBDT-GP} & \textbf{Decision Tree} & \textbf{Random Forest} & \textbf{Gaussian NB}\\
            \midrule
            Accuracy & 0.9840  & 0.9847 & \textbf{0.9868} & 0.7585 & 0.9838 & 0.6862\\ 
            F1-Score & 0.9840 & 0.9845 & \textbf{0.9867} & 0.7085 & 0.9838 & 0.6674\\
            Precision & 0.9857 & 0.9872 & \textbf{0.9884} & 0.6869 & 0.9851 & 0.7025\\
            Recall & 0.9840 & 0.9847 & \textbf{0.9868} & 0.7672 & 0.9841 & 0.6867\\
            Avg. Terminals & 3.70 & 3.74 & \textbf{3.59} & --- & --- & ---\\
            \bottomrule
        \end{tabular}
    \label{tab:auslanmetrics}
\end{table*}

\begin{figure}[!ht]
    \centering
    \resizebox{1.0\columnwidth}{!}
    {
        \input{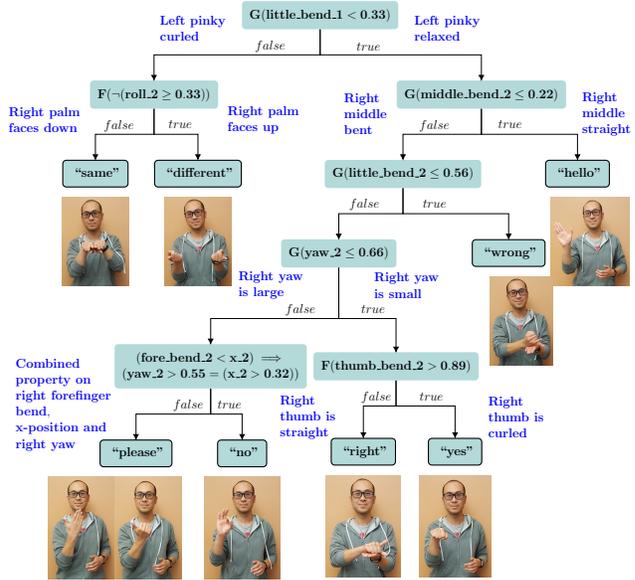}
    }
    \caption{GBDT categorization of the Auslan dataset.}
    \label{fig:auslan}

\end{figure}

The strength of \ac{gbdt} lies in its ability to categorize and explain data.  \ac{gbdt} not only provides interpretable expressions about the data's distinguishing properties but also a hierarchical organization of the properties.  \Cref{fig:auslan} shows the resulting \ac{gbdt} from the best-performing trial.  We compare the learned model to the videos of the signed words to intuitively verify correctness\footnote{\url{https://www.youtube.com/watch?v=_5NbYyUlcHU}}.  
At the root node, \ac{gbdt} partitions \textit{same} and \textit{different} from the other words identifying the fully bent left pinky as the distinguishing property.  Subsequently, \textit{different} is distinguished from \textit{same} by looking at whether the right palm is ever facing upwards.  Of the remaining words, \textit{hello} is isolated using the fully straight middle finger and then \textit{wrong} is identified using the straight right pinky.  The remaining four words are grouped \textit{right} with \textit{yes} and \textit{please} with \textit{no} using the yaw position of the right hand.  The word \textit{yes} is distinguished from \textit{right} by the bent right thumb.  Lastly, \textit{no} is distinguished from \textit{please} using the combined property on right forefinger bend, right yaw angle, and $x$-position of the right hand.

\subsection{Collision Avoidance Application.}
\label{sec:application}

Airborne collision avoidance systems are mandated worldwide on all large transport and cargo aircraft to help prevent mid-air collisions.  They monitor the airspace around an aircraft and issue alerts, called \acp{ra}, if a conflict is detected.  \Iac{ra} recommends a safety action to the pilot, for example, instructing the pilot to climb at 1500 feet per minute.  To address the growing needs of the national airspace, the \ac{faa} is currently developing and testing a new aircraft collision avoidance system, called the next-generation \acf{acasx}. \ac{acasx} promises a number of improvements over current systems including a reduction in collision risk while simultaneously reducing the number of unnecessary alerts \cite{Kochenderfer2012next}.  

One of the primary safety metrics of airborne collision avoidance systems is the likelihood of \ac{nmac}, defined as two aircraft coming closer than 500 feet horizontally and 100 feet vertically. Efficient algorithms have been developed to generate large datasets of \ac{nmac} and non-\ac{nmac} instances in simulation \cite{Lee2015}. However, while it is straightforward to observe that \iac{nmac} has occurred, discovering and categorizing relevant properties of \acp{nmac} is challenging.  

We apply \ac{gbdt} to analyze simulated aircraft encounters to discover the most predictive properties of \acp{nmac} and categorize encounters accordingly.  The results of our study are used to help the \ac{acasx} development team better understand the \acp{nmac} for validating safety and informing development.

\textbf{Dataset.} We analyze a dataset that contains simulated two-aircraft mid-air encounters \cite{Lee2015}.  The dataset contains 10,000 encounters with 863 \acp{nmac} and 9,137 non-\acp{nmac}.  The class imbalance is due to the rarity of \acp{nmac} and the difficulty in generating \ac{nmac} encounters.  Each encounter has 77 attributes collected at \SI{1}{\hertz} for 50 time steps.  The attributes include numeric, categorical, and Boolean types representing the state of the aircraft, pilot commands, and the state and output of the collision avoidance system for each aircraft.  The data was generated during the stress testing of \ac{acasx} prototype version 0.9.8.

%

\textbf{Grammar.} We craft a custom \ac{cfg} for the \ac{acasx} dataset, building on the one presented in \Cref{fig:gbdtexample}. We include temporal logic operators \textit{eventually} \text{F} and \textit{globally} \text{G}; elementwise logical operators \textit{conjunct} $\wedge$, \textit{disjunct} $\vee$, \textit{negation} $\neg$, and \textit{implies} $\implies$; comparison operators \textit{less than} $<$, \textit{less than or equal to} $\leq$, \textit{greater than} $>$, \textit{greater than or equal to} $\geq$, and \textit{equal} $=$; mathematical functions \textit{absolute value} $|x|$, \textit{difference} $-$, and \textit{sign} $\textsc{sign}$; and \textit{count} $\textsc{count}$ (which returns the number of true values in a Boolean vector).  

In addition to dividing the attributes by data type, the \ac{acasx} grammar further subdivides the attributes by their physical representations.  This enables comparison of attributes with constant values that have appropriate scale and resolution.  For example, even though aircraft heading and vertical rate are both real-valued, aircraft heading should be compared to values between \num{-180}\si{\degree} and \num{180}\si{\degree}, whereas vertical rate should be compared to values between \num{-80} and \num{80} feet per second.  

\subsubsection{Results.} We performed 20 random trials for each algorithm and report the mean results in Table \ref{tab:acasxmetrics}.  \ac{gbdt}, decision tree, and random forest were trained to a maximum depth $d=4$.  Random forest used 64 trees.  \ac{gbdt}-GP again performed the best overall.  The temporal logic rules in \ac{gbdt} are able to more effectively capture the underlying patterns and are robust to temporal variations.  Random forest performed very poorly at predicting \ac{nmac}.  The reason is that random forest randomly subsamples attributes, which works poorly when there are many attributes but only a small subset of attributes are predictive.   

\begin{table*}[!ht]
    \centering
    \footnotesize
    \caption{Algorithm performance metrics on ACAS X dataset. Best value for each metric is in bold.} 
        \begin{tabular}{@{}lcccccc}
            \toprule
            \textbf{} & \textbf{GBDT-MC} & \textbf{GBDT-GE} & \textbf{GBDT-GP} & \textbf{Decision Tree} & \textbf{Random Forest} & \textbf{Gaussian NB}\\
            \midrule
            Accuracy & 0.9577  & 0.9583 & \textbf{0.9587} & 0.9378 & 0.9251& 0.8705\\ 
            F1-Score & 0.8765 & 0.8768 & \textbf{0.8779} & 0.8038 & 0.6130 & 0.7126\\
            Precision & 0.8505 & 0.8546 & 0.8553 & 0.8025 & \textbf{0.8961} & 0.6799\\
            Recall & \textbf{0.9102} & 0.9050 & 0.9053 & 0.8083 & 0.5778 & 0.8187\\
            Avg. Terminals & 5.54 & 5.56 & \textbf{5.38} & --- & --- & ---\\
            \bottomrule
        \end{tabular}
    \label{tab:acasxmetrics}
\end{table*}

\begin{figure}[!ht]
    \centering
    \resizebox{1.0\columnwidth}{!}
    {
        \includegraphics[width=\columnwidth]{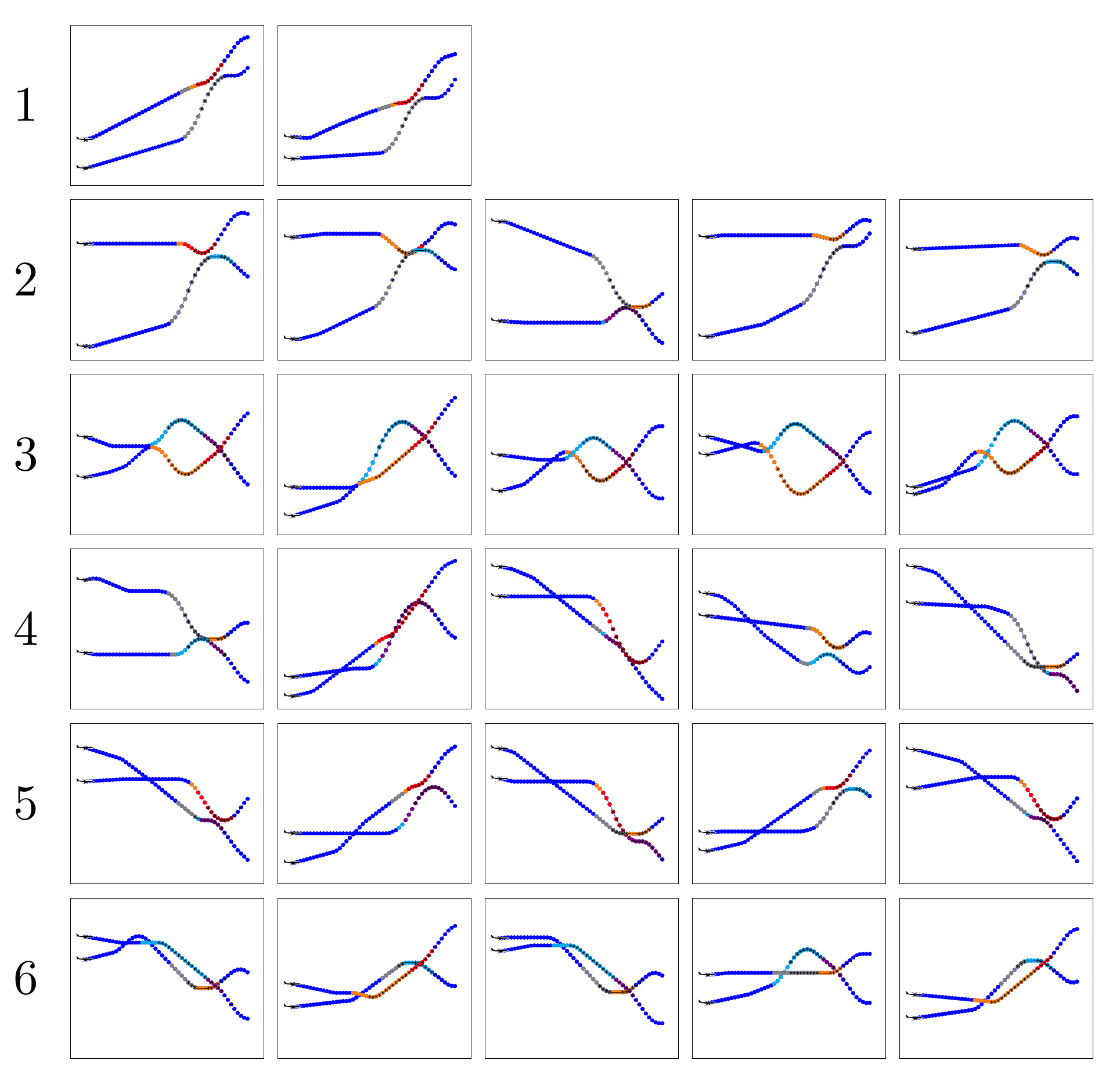}
    }
    \caption{Visual overview of categorized ACAS X encounter data. Each row is a category.}
    \label{fig:acasxcollage}

\end{figure}

\Cref{fig:acasxcollage} shows a visual overview of a categorization from one of the resulting trees.  The figure shows plots of altitude versus time, which, while cannot fully capture the high-dimensionality of the encounter data, is generally most informative since \ac{acasx} issues \acp{ra} only in the vertical direction.  Since we are mainly interested in categorizing \acp{nmac}, we consider only the six \ac{nmac} categories out of the 16 total categories produced by the tree.  \Cref{fig:acasxcollage} shows the first five encounters for each of the six \ac{nmac} categories, where each row is a separate category.  The categories are labeled in ascending order starting at category 1 at the top row.  The first row has only two plots because category 1 only contained two encounters.  
We describe categories 1, 5, and 6 as follows. 

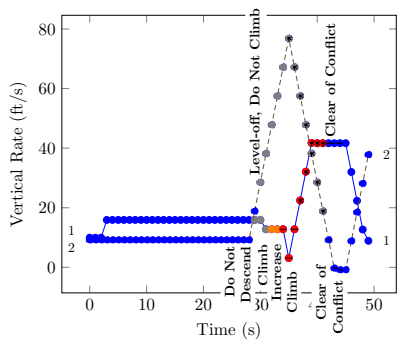
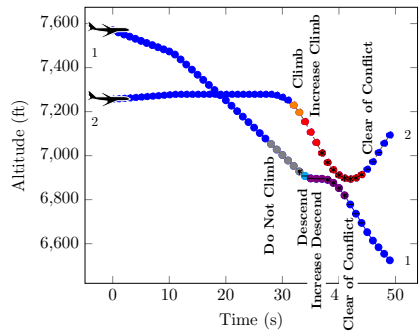
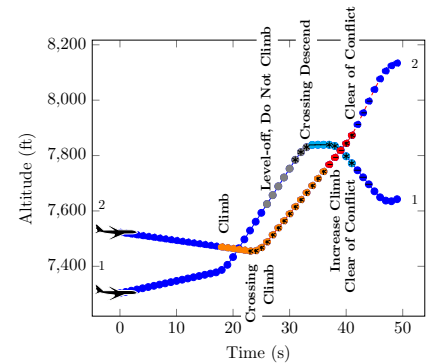
\begin{figure*}[ht!]
    \begin{subfigure}[b]{0.315\textwidth}
        \resizebox{\textwidth}{!}
        {
        \begin{tikzpicture}[]
\begin{groupplot}[group style={horizontal sep = 2.2cm, vertical sep = 2.2cm, group size=1 by 1}]
\nextgroupplot [clip mode=individual, title = {}, xlabel = {Time (s)}, ylabel = {Vertical Rate (ft/s)}]\node at (axis cs:0.0, 12.5) [left=2.0mm,scale=0.8,rotate=0] {1};
\node at (axis cs:49.0, 8.904895833) [right=2.0mm,scale=0.8,rotate=0] {1};
\node at (axis cs:0.0, 6.5) [left=2.0mm,scale=0.8,rotate=0] {2};
\node at (axis cs:49.0, 37.839383916) [right=2.0mm,scale=0.8,rotate=0] {2};

\node at (axis cs:25.0, 11.5) [rotate=90,scale=0.8,xshift=-2mm,left=2mm,yshift=-2mm,fill=white,rectangle,rounded corners=3pt] {\textbf{Do Not}};
\node at (axis cs:27.8, 11.5) [rotate=90,scale=0.8,xshift=-2mm,left=2mm,yshift=-2mm,fill=white,rectangle,rounded corners=3pt] {\textbf{ Descend}};
\node at (axis cs:30.8, 12.0) [rotate=90,scale=0.8,xshift=-2mm,left=2mm,yshift=-2mm,fill=white,rectangle,rounded corners=3pt] {\textbf{Climb}};
\node at (axis cs:33.2, 12.0) [rotate=90,scale=0.8,xshift=-2mm,left=2mm,yshift=-2mm,fill=white,rectangle,rounded corners=3pt] {\textbf{Increase}};
\node at (axis cs:35.9, 3.0) [rotate=90,scale=0.8,xshift=-2mm,left=2mm,yshift=-2mm,fill=white,rectangle,rounded corners=3pt] {\textbf{Climb}};
\node at (axis cs:42.0, 42.0) [rotate=90,scale=0.8,yshift=2mm,right=2mm,fill=white,rectangle,rounded corners=3pt] {\textbf{Clear of Conflict}};
\node at (axis cs:28.2, 29.5) [rotate=90,scale=0.8,right=2mm,yshift=1mm,fill=white,rectangle,rounded corners=3pt] {\textbf{Level-off, Do Not Climb}};
\node at (axis cs:41.0, 5.5) [rotate=90,scale=0.8,left=2mm,xshift=-2mm,yshift=-2mm,fill=white,rectangle,rounded corners=3pt] {\textbf{Clear of}};
\node at (axis cs:44.0, 0.0) [rotate=90,scale=0.8,left=2mm,xshift=-2mm,yshift=-2mm,fill=white,rectangle,rounded corners=3pt] {\textbf{Conflict}};

\addplot+ [mark = {*}, mark options={color=blue}]coordinates {
(0.0, 9.957122318)
(1.0, 9.957122286)
(2.0, 9.957122286)
(3.0, 15.911410583)
(4.0, 15.911410583)
(5.0, 15.911410583)
(6.0, 15.911410583)
(7.0, 15.911410582)
(8.0, 15.911410582)
(9.0, 15.911410582)
(10.0, 15.911410582)
(11.0, 15.911410582)
(12.0, 15.911410582)
(13.0, 15.911410582)
(14.0, 15.911410582)
(15.0, 15.911410582)
(16.0, 15.911410582)
(17.0, 15.911410582)
(18.0, 15.911410582)
(19.0, 15.911410582)
(20.0, 15.911410582)
(21.0, 15.911410582)
(22.0, 15.911410582)
(23.0, 15.911410582)
(24.0, 15.911410581)
(25.0, 15.911410581)
(26.0, 15.911410581)
(27.0, 15.911410581)
(28.0, 15.911410581)
(29.0, 15.911410581)
(30.0, 15.911410581)
(31.0, 12.786543284)
(32.0, 12.786543284)
(33.0, 12.786543284)
(34.0, 12.786543284)
(35.0, 3.126947366)
(36.0, 12.786233458)
(37.0, 22.445329943)
(38.0, 32.104235152)
(39.0, 41.76294695)
(40.0, 41.666673419)
(41.0, 41.666673354)
(42.0, 41.66667329)
(43.0, 41.666673225)
(44.0, 41.66667316)
(45.0, 41.666673095)
(46.0, 32.006509147)
(47.0, 22.346536114)
(48.0, 12.686753602)
(49.0, 8.904895833)
};
\addplot+ [mark options={color=gray}, mark=*, only marks = {true}]coordinates {
(29.0, 15.911410581)
(30.0, 15.911410581)
(31.0, 12.786543284)
};
\addplot+ [mark options={color=orange}, mark=*, only marks = {true}]coordinates {
(32.0, 12.786543284)
(33.0, 12.786543284)
};
\addplot+ [mark options={color=red}, mark=*, only marks = {true}]coordinates {
(34.0, 12.786543284)
(35.0, 3.126947366)
(36.0, 12.786233458)
(37.0, 22.445329943)
(38.0, 32.104235152)
(39.0, 41.76294695)
(40.0, 41.666673419)
(41.0, 41.666673354)
};
\addplot+ [mark options={color=black}, mark=-, only marks = {true}]coordinates {
(34.0, 12.786543284)
(35.0, 3.126947366)
(36.0, 12.786233458)
(42.0, 41.66667329)
(43.0, 41.666673225)
(44.0, 41.66667316)
};
\addplot+ [mark options={color=black}, mark=asterisk, only marks = {true}]coordinates {
(37.0, 22.445329943)
(38.0, 32.104235152)
(39.0, 41.76294695)
(40.0, 41.666673419)
(41.0, 41.666673354)
};
\addplot+ [mark = {*}, mark options={color=blue}]coordinates {
(0.0, 9.213245301)
(1.0, 9.213245147)
(2.0, 9.213245147)
(3.0, 9.213245147)
(4.0, 9.213245148)
(5.0, 9.213245148)
(6.0, 9.213245149)
(7.0, 9.213245149)
(8.0, 9.213245149)
(9.0, 9.21324515)
(10.0, 9.21324515)
(11.0, 9.213245151)
(12.0, 9.213245151)
(13.0, 9.213245151)
(14.0, 9.213245152)
(15.0, 9.213245152)
(16.0, 9.213245152)
(17.0, 9.213245153)
(18.0, 9.213245153)
(19.0, 9.213245154)
(20.0, 9.213245154)
(21.0, 9.213245154)
(22.0, 9.213245155)
(23.0, 9.213245155)
(24.0, 9.213245155)
(25.0, 9.213245156)
(26.0, 9.213245156)
(27.0, 9.213245157)
(28.0, 9.213245157)
(29.0, 18.87425645)
(30.0, 28.535131483)
(31.0, 38.195871062)
(32.0, 47.856475765)
(33.0, 57.516945949)
(34.0, 67.177281746)
(35.0, 76.837483067)
(36.0, 67.175250704)
(37.0, 57.513166401)
(38.0, 47.851227225)
(39.0, 38.189430528)
(40.0, 28.527773933)
(41.0, 18.866255324)
(42.0, 9.204872834)
(43.0, -0.294705309)
(44.0, -0.804349226)
(45.0, -0.804349228)
(46.0, 8.856780172)
(47.0, 18.517778209)
(48.0, 28.178645866)
(49.0, 37.839383916)
};
\addplot+ [mark options={color=gray}, mark=*, only marks = {true}]coordinates {
(30.0, 28.535131483)
(31.0, 38.195871062)
(32.0, 47.856475765)
(33.0, 57.516945949)
(34.0, 67.177281746)
(35.0, 76.837483067)
(36.0, 67.175250704)
(37.0, 57.513166401)
(38.0, 47.851227225)
(39.0, 38.189430528)
(40.0, 28.527773933)
(41.0, 18.866255324)
};
\addplot+ [mark options={color=black}, mark=-, only marks = {true}]coordinates {
(42.0, 9.204872834)
(43.0, -0.294705309)
(44.0, -0.804349226)
};
\addplot+ [mark options={color=black}, mark=asterisk, only marks = {true}]coordinates {
(35.0, 76.837483067)
(36.0, 67.175250704)
(37.0, 57.513166401)
(38.0, 47.851227225)
(39.0, 38.189430528)
(40.0, 28.527773933)
(41.0, 18.866255324)
};
\end{groupplot}

\end{tikzpicture}
        }
        \caption{Category 1: Vertical\\ maneuvering just before NMAC.}
        \label{fig:cat1}
    \end{subfigure}%
    \begin{subfigure}[b]{0.33\textwidth}
        \resizebox{\textwidth}{!}
        {
            \begin{tikzpicture}[]
\begin{groupplot}[group style={horizontal sep = 2.2cm, vertical sep = 2.2cm, group size=1 by 1}]
\nextgroupplot [clip mode=individual, title = {}, xlabel = {Time (s)}, clip=false,clip mode=individual, ylabel = {Altitude (ft)}]
\node at (axis cs:28.0, 7052.633239238) [rotate=90,scale=0.8,xshift=-2mm,yshift=-2mm,left=2mm,fill=white,rectangle,rounded corners=3pt] {\textbf{Do Not Climb}};
\node at (axis cs:32.5, 6907.551186627) [rotate=90,scale=0.8,xshift=-2mm,yshift=-4mm,left=2mm,fill=white,rectangle,rounded corners=3pt] {\textbf{Descend}};
\node at (axis cs:35.7, 6896.652544636) [rotate=90,scale=0.8,xshift=-2mm,yshift=-3mm,left=2mm,fill=white,rectangle,rounded corners=3pt] {\textbf{Increase Descend}};
\node at (axis cs:42.0, 6778.198316878) [rotate=90,scale=0.8,left=2mm,xshift=-2mm,yshift=-2mm,fill=white,rectangle,rounded corners=3pt] {\textbf{Clear of Conflict}};
\node at (axis cs:31.0, 7278.992186936) [rotate=90,scale=0.8,right=2mm,fill=white,rectangle,rounded corners=3pt] {\textbf{Climb}};
\node at (axis cs:34.0, 7154.161694736) [rotate=90,scale=0.8,right=2mm,fill=white,rectangle,rounded corners=3pt] {\textbf{Increase Climb}};
\node at (axis cs:44.7, 6980.0846817) [rotate=90,scale=0.8,yshift=2mm,right=2mm,fill=white,rectangle,rounded corners=3pt] {\textbf{Clear of Conflict}};
\addplot+ [mark = {*}, mark options={color=blue}]coordinates {
(0.0, 7569.522583818)
(1.0, 7559.899380089)
(2.0, 7550.276176368)
(3.0, 7540.652972647)
(4.0, 7531.029768927)
(5.0, 7521.406565207)
(6.0, 7511.783361486)
(7.0, 7502.160157767)
(8.0, 7492.536954047)
(9.0, 7482.913750327)
(10.0, 7473.290546608)
(11.0, 7459.320875489)
(12.0, 7437.120529132)
(13.0, 7413.933549673)
(14.0, 7390.746570214)
(15.0, 7367.559590756)
(16.0, 7344.372611298)
(17.0, 7321.18563184)
(18.0, 7297.998652383)
(19.0, 7274.811672927)
(20.0, 7251.624693471)
(21.0, 7226.966877792)
(22.0, 7202.062072283)
(23.0, 7177.157266774)
(24.0, 7152.252461266)
(25.0, 7127.347655758)
(26.0, 7102.442850251)
(27.0, 7077.538044745)
(28.0, 7052.633239238)
(29.0, 7027.728433733)
(30.0, 7002.823628227)
(31.0, 6977.918822723)
(32.0, 6953.014017218)
(33.0, 6928.109211715)
(34.0, 6907.551186627)
(35.0, 6896.652544636)
(36.0, 6894.74977102)
(37.0, 6894.74977102)
(38.0, 6890.403228517)
(39.0, 6876.3978341)
(40.0, 6852.733796905)
(41.0, 6819.41132873)
(42.0, 6778.198316878)
(43.0, 6736.14529069)
(44.0, 6694.188845304)
(45.0, 6652.232400568)
(46.0, 6613.946842216)
(47.0, 6583.575590366)
(48.0, 6553.972639324)
(49.0, 6524.369688283)
};
\addplot+ [mark options={color=gray}, mark=*, only marks = {true}]coordinates {
(28.0, 7052.633239238)
(29.0, 7027.728433733)
(30.0, 7002.823628227)
(31.0, 6977.918822723)
(32.0, 6953.014017218)
(33.0, 6928.109211715)
};
\addplot+ [mark options={color=cyan}, mark=*, only marks = {true}]coordinates {
(34.0, 6907.551186627)
};
\addplot+ [mark options={color=violet}, mark=*, only marks = {true}]coordinates {
(35.0, 6896.652544636)
(36.0, 6894.74977102)
(37.0, 6894.74977102)
(38.0, 6890.403228517)
(39.0, 6876.3978341)
(40.0, 6852.733796905)
(41.0, 6819.41132873)
};
\addplot+ [mark options={color=black}, mark=-, only marks = {true}]coordinates {
(34.0, 6907.551186627)
(35.0, 6896.652544636)
(36.0, 6894.74977102)
(37.0, 6894.74977102)
(42.0, 6778.198316878)
(43.0, 6736.14529069)
(44.0, 6694.188845304)
};
\addplot+ [mark options={color=black}, mark=asterisk, only marks = {true}]coordinates {
(33.0, 6928.109211715)
(38.0, 6890.403228517)
(39.0, 6876.3978341)
(40.0, 6852.733796905)
(41.0, 6819.41132873)
};
\addplot+ [mark = {*}, mark options={color=blue}]coordinates {
(0.0, 7255.979751824)
(1.0, 7257.992775213)
(2.0, 7260.005798592)
(3.0, 7262.018821971)
(4.0, 7264.031845351)
(5.0, 7266.044868731)
(6.0, 7268.057892112)
(7.0, 7270.070915493)
(8.0, 7272.083938874)
(9.0, 7274.096962256)
(10.0, 7276.109985639)
(11.0, 7278.123009022)
(12.0, 7278.437049145)
(13.0, 7278.437049145)
(14.0, 7278.437049145)
(15.0, 7278.437049145)
(16.0, 7278.437049145)
(17.0, 7278.437049145)
(18.0, 7278.437049145)
(19.0, 7278.437049145)
(20.0, 7278.437049145)
(21.0, 7278.437049145)
(22.0, 7278.437049145)
(23.0, 7278.437049145)
(24.0, 7278.437049145)
(25.0, 7278.437049145)
(26.0, 7278.437049145)
(27.0, 7278.437049145)
(28.0, 7278.437049145)
(29.0, 7274.089136238)
(30.0, 7265.175988265)
(31.0, 7251.914991857)
(32.0, 7228.992186936)
(33.0, 7196.407709709)
(34.0, 7154.161694736)
(35.0, 7106.761093529)
(36.0, 7059.360505095)
(37.0, 7011.959916652)
(38.0, 6968.907521988)
(39.0, 6935.517688525)
(40.0, 6911.790273792)
(41.0, 6897.725138191)
(42.0, 6893.322144743)
(43.0, 6898.58115884)
(44.0, 6913.50204801)
(45.0, 6938.0846817)
(46.0, 6972.328931058)
(47.0, 7014.021575352)
(48.0, 7056.171092189)
(49.0, 7093.48941197)
};
\addplot+ [mark options={color=orange}, mark=*, only marks = {true}]coordinates {
(32.0, 7228.992186936)
(33.0, 7196.407709709)
};
\addplot+ [mark options={color=red}, mark=*, only marks = {true}]coordinates {
(34.0, 7154.161694736)
(35.0, 7106.761093529)
(36.0, 7059.360505095)
(37.0, 7011.959916652)
(38.0, 6968.907521988)
(39.0, 6935.517688525)
(40.0, 6911.790273792)
(41.0, 6897.725138191)
(42.0, 6893.322144743)
(43.0, 6898.58115884)
(44.0, 6913.50204801)
};
\addplot+ [mark options={color=black}, mark=-, only marks = {true}]coordinates {
(45.0, 6938.0846817)
(46.0, 6972.328931058)
(47.0, 7014.021575352)
};
\addplot+ [mark options={color=black}, mark=asterisk, only marks = {true}]coordinates {
(37.0, 7011.959916652)
(38.0, 6968.907521988)
(39.0, 6935.517688525)
(40.0, 6911.790273792)
(41.0, 6897.725138191)
(42.0, 6893.322144743)
(43.0, 6898.58115884)
(44.0, 6913.50204801)
};
\node at (axis cs:0.0, 7569.522583818) [aircraft side,draw=white,thin,fill=black,minimum width=0.9cm,rotate=0] {};
\node at (axis cs:0.0, 7255.979751824) [aircraft side,draw=white,thin,fill=black,minimum width=0.9cm,rotate=0] {};
\node at (axis cs:0.0, 7469.522583818) [left=2.0mm,scale=0.8,rotate=0] {1};
\node at (axis cs:49.0, 6524.369688283) [right=2.0mm,scale=0.8,rotate=0] {1};
\node at (axis cs:0.0, 7155.979751824) [left=2.0mm,scale=0.8,rotate=0] {2};
\node at (axis cs:49.0, 7093.48941197) [right=2.0mm,scale=0.8,rotate=0] {2};
\end{groupplot}

\end{tikzpicture}
        }
        \caption{Category 5: Maneuvering\\ following altitude crossing.}
        \label{fig:cat5}
    \end{subfigure}%
    \begin{subfigure}[b]{0.33\textwidth}
        \resizebox{\textwidth}{!}
        {
            \begin{tikzpicture}[]
\begin{groupplot}[group style={horizontal sep = 2.2cm, vertical sep = 2.2cm, group size=1 by 1}]
\nextgroupplot [clip mode=individual, title = {}, xlabel = {Time (s)}, clip=false,clip mode=individual, ylabel = {Altitude (ft)}]
\node at (axis cs:25.7, 7624.995645494) [rotate=90,scale=0.8,xshift=2mm,yshift=2mm,right=2mm,fill=white,rectangle,rounded corners=3pt] {\textbf{Level-off, Do Not Climb}};
\node at (axis cs:34.0, 7838.426688072) [rotate=90,scale=0.8,yshift=4mm,right=2mm,fill=white,rectangle,rounded corners=3pt] {\textbf{Crossing Descend}};
\node at (axis cs:41.7, 7746.874061558) [rotate=90,scale=0.8,left=2mm,xshift=-1mm,yshift=-1mm,fill=white,rectangle,rounded corners=3pt] {\textbf{Clear of Conflict}};
\node at (axis cs:18.0, 7469.651505886) [rotate=90,scale=0.8,right=2mm,yshift=2mm,xshift=2mm,fill=white,rectangle,rounded corners=3pt] {\textbf{Climb}};
\node at (axis cs:25.0, 7467.628346487) [rotate=90,scale=0.8,left=2mm,xshift=-3mm,fill=white,rectangle,rounded corners=3pt] {\textbf{Crossing}};
\node at (axis cs:28.0, 7467.628346487) [rotate=90,scale=0.8,left=2mm,fill=white,yshift=0mm,xshift=-3mm,rectangle,rounded corners=3pt] {\textbf{Climb}};

\node at (axis cs:37.0, 7767.800367937) [rotate=90,scale=0.8,yshift=-4mm,left=2mm,fill=white,rectangle,rounded corners=3pt] {\textbf{Increase Climb}};
\node at (axis cs:41.7, 7912.194014643) [rotate=90,scale=0.8,right=2mm,yshift=4mm,fill=white,rectangle,rounded corners=3pt] {\textbf{Clear of Conflict}};
\addplot+ [mark = {*}, mark options={color=blue}]coordinates {
(0.0, 7303.050878531)
(1.0, 7307.470416203)
(2.0, 7311.889953872)
(3.0, 7316.30949154)
(4.0, 7320.729029209)
(5.0, 7325.148566878)
(6.0, 7329.568104546)
(7.0, 7333.987642215)
(8.0, 7338.407179883)
(9.0, 7342.826717552)
(10.0, 7347.24625522)
(11.0, 7351.665792888)
(12.0, 7356.085330556)
(13.0, 7360.504868225)
(14.0, 7364.924405893)
(15.0, 7369.343943561)
(16.0, 7373.763481229)
(17.0, 7378.183018897)
(18.0, 7386.94913558)
(19.0, 7405.37422464)
(20.0, 7433.408223756)
(21.0, 7465.339460713)
(22.0, 7497.27069767)
(23.0, 7529.201934627)
(24.0, 7561.133171583)
(25.0, 7593.064408539)
(26.0, 7624.995645494)
(27.0, 7656.926882449)
(28.0, 7688.858119403)
(29.0, 7720.789356358)
(30.0, 7752.720593311)
(31.0, 7784.651830265)
(32.0, 7812.236245223)
(33.0, 7830.161148765)
(34.0, 7838.426688072)
(35.0, 7838.953399419)
(36.0, 7838.953399419)
(37.0, 7838.953399419)
(38.0, 7834.606797939)
(39.0, 7820.601177112)
(40.0, 7797.453618345)
(41.0, 7772.163837686)
(42.0, 7746.874061558)
(43.0, 7721.487697878)
(44.0, 7696.197917344)
(45.0, 7670.908141342)
(46.0, 7649.289046432)
(47.0, 7637.329368367)
(48.0, 7635.028956118)
(49.0, 7642.387657821)
};
\addplot+ [mark options={color=gray}, mark=*, only marks = {true}]coordinates {
(26.0, 7624.995645494)
(27.0, 7656.926882449)
(28.0, 7688.858119403)
(29.0, 7720.789356358)
(30.0, 7752.720593311)
(31.0, 7784.651830265)
(32.0, 7812.236245223)
(33.0, 7830.161148765)
};
\addplot+ [mark options={color=cyan}, mark=*, only marks = {true}]coordinates {
(34.0, 7838.426688072)
(35.0, 7838.953399419)
(36.0, 7838.953399419)
(37.0, 7838.953399419)
(38.0, 7834.606797939)
(39.0, 7820.601177112)
(40.0, 7797.453618345)
(41.0, 7772.163837686)
};
\addplot+ [mark options={color=black}, mark=-, only marks = {true}]coordinates {
(34.0, 7838.426688072)
(35.0, 7838.953399419)
(36.0, 7838.953399419)
(42.0, 7746.874061558)
(43.0, 7721.487697878)
(44.0, 7696.197917344)
};
\addplot+ [mark options={color=black}, mark=asterisk, only marks = {true}]coordinates {
(31.0, 7784.651830265)
(32.0, 7812.236245223)
(33.0, 7830.161148765)
(37.0, 7838.953399419)
(38.0, 7834.606797939)
(39.0, 7820.601177112)
(40.0, 7797.453618345)
(41.0, 7772.163837686)
};
\addplot+ [mark = {*}, mark options={color=blue}]coordinates {
(0.0, 7522.060778049)
(1.0, 7519.149151808)
(2.0, 7516.237525585)
(3.0, 7513.32589936)
(4.0, 7510.414273135)
(5.0, 7507.502646908)
(6.0, 7504.591020681)
(7.0, 7501.679394453)
(8.0, 7498.767768224)
(9.0, 7495.856141994)
(10.0, 7492.944515763)
(11.0, 7490.032889531)
(12.0, 7487.121263299)
(13.0, 7484.209637066)
(14.0, 7481.298010831)
(15.0, 7478.386384596)
(16.0, 7475.47475836)
(17.0, 7472.563132124)
(18.0, 7469.651505886)
(19.0, 7466.739879648)
(20.0, 7463.828253409)
(21.0, 7460.916627169)
(22.0, 7458.005000928)
(23.0, 7455.093374687)
(24.0, 7456.529780788)
(25.0, 7467.628346487)
(26.0, 7488.378044721)
(27.0, 7513.780112968)
(28.0, 7539.182136291)
(29.0, 7564.584160104)
(30.0, 7589.986184406)
(31.0, 7615.388209193)
(32.0, 7640.790234461)
(33.0, 7666.192260209)
(34.0, 7691.594286434)
(35.0, 7716.996313132)
(36.0, 7742.3983403)
(37.0, 7767.800367937)
(38.0, 7793.202396039)
(39.0, 7818.604424604)
(40.0, 7844.006453629)
(41.0, 7873.353977934)
(42.0, 7912.194014643)
(43.0, 7954.343487384)
(44.0, 7996.492951796)
(45.0, 8038.642417032)
(46.0, 8076.92661332)
(47.0, 8105.547966351)
(48.0, 8124.506677797)
(49.0, 8133.802944813)
};
\addplot+ [mark options={color=orange}, mark=*, only marks = {true}]coordinates {
(18.0, 7469.651505886)
(19.0, 7466.739879648)
(20.0, 7463.828253409)
(21.0, 7460.916627169)
(22.0, 7458.005000928)
(23.0, 7455.093374687)
(24.0, 7456.529780788)
(25.0, 7467.628346487)
(26.0, 7488.378044721)
(27.0, 7513.780112968)
(28.0, 7539.182136291)
(29.0, 7564.584160104)
(30.0, 7589.986184406)
(31.0, 7615.388209193)
(32.0, 7640.790234461)
(33.0, 7666.192260209)
(34.0, 7691.594286434)
(35.0, 7716.996313132)
(36.0, 7742.3983403)
};
\addplot+ [mark options={color=red}, mark=*, only marks = {true}]coordinates {
(37.0, 7767.800367937)
(38.0, 7793.202396039)
(39.0, 7818.604424604)
(40.0, 7844.006453629)
(41.0, 7873.353977934)
};
\addplot+ [mark options={color=black}, mark=-, only marks = {true}]coordinates {
(37.0, 7767.800367937)
(38.0, 7793.202396039)
(39.0, 7818.604424604)
(42.0, 7912.194014643)
(43.0, 7954.343487384)
(44.0, 7996.492951796)
};
\addplot+ [mark options={color=black}, mark=asterisk, only marks = {true}]coordinates {
(23.0, 7455.093374687)
(24.0, 7456.529780788)
(25.0, 7467.628346487)
(26.0, 7488.378044721)
(27.0, 7513.780112968)
(28.0, 7539.182136291)
(29.0, 7564.584160104)
(30.0, 7589.986184406)
(31.0, 7615.388209193)
(32.0, 7640.790234461)
(33.0, 7666.192260209)
(34.0, 7691.594286434)
(35.0, 7716.996313132)
(36.0, 7742.3983403)
(40.0, 7844.006453629)
(41.0, 7873.353977934)
};
\node at (axis cs:0.0, 7303.050878531) [aircraft side,draw=white,thin,fill=black,minimum width=0.9cm,rotate=0] {};
\node at (axis cs:0.0, 7522.060778049) [aircraft side,draw=white,thin,fill=black,minimum width=0.9cm,rotate=0] {};
\node at (axis cs:0.0, 7403.050878531) [left=2.0mm,scale=0.8,rotate=0] {1};
\node at (axis cs:49.0, 7642.387657821) [right=2.0mm,scale=0.8,rotate=0] {1};
\node at (axis cs:0.0, 7622.060778049) [left=2.0mm,scale=0.8,rotate=0] {2};
\node at (axis cs:49.0, 8133.802944813) [right=2.0mm,scale=0.8,rotate=0] {2};
\end{groupplot}

\end{tikzpicture}
        }
        \caption{Category 6: Altitude crossing\\ during pilot response delay.}
        \label{fig:cat6}
    \end{subfigure}
    \caption{NMAC examples. Textual labels indicate issued \acp{ra}.  Aircraft numbers are labeled at the beginning and end of a plot.  An asterisk marker indicates that the pilot is following \iac{ra} while a dash marker indicates a pilot response delay.  Blue marker indicates no RA.  Other colors indicate an active RA.} 
    \label{fig:cats}
\end{figure*}

\textbf{Category 1.} In this category, the two aircraft maintain altitude separation for most of the encounter, then one aircraft accelerates rapidly toward the other aircraft as they approach.  The aggressive maneuvering causes vertical separation to be lost rapidly and \iac{nmac} results.  \Cref{fig:cat1} shows the spike in climb rate at 34 seconds, only 5 seconds before \ac{nmac}.  The large spike in vertical rate near \ac{nmac} is characteristic of encounters in this category.  


Aggressive last-minute maneuvering is known to be problematic.  With the pilot's five-second response time, there is insufficient time remaining for the collision avoidance system to resolve the conflict.  It is unlikely that any reasonable collision avoidance system would be able to resolve such \acp{nmac}.  Fortunately, these encounters are extremely rare outside of simulation since pilots do not typically maneuver so aggressively in operations. 

\textbf{Category 5.} In these encounters, the aircraft cross in altitude early in the encounter without active advisories, then maneuver back toward each other to cause \iac{nmac}.  Since the aircraft are predicted to cross safely and appear to vertically diverge following the crossing, the collision avoidance system witholds \iac{ra} to allow the encounter to resolve naturally and reduce the number of unnecessary alerts.  However, after crossing, the aircraft change course and maneuver toward each other, which results in \iac{nmac}.  Because the aircraft are already close in altitude, the vertical maneuvering in these encounters is less aggressive than those in previous categories.  Due to the late maneuvering and the pilot response delay, pilot 2 does not start to comply with the issued \ac{ra} until within five seconds of \ac{nmac}.  \Cref{fig:cat5} shows the vertical profile of an encounter in this category where the aircraft cross in altitude at 19 seconds and then maneuver to \ac{nmac} at 38 seconds.


\textbf{Category 6.} In these encounters, the aircraft receive initial \acp{ra} at different times and the aircraft cross in altitude during the pilot response delay of the first \ac{ra}.  As the pilots start responding to their advisories, the maneuvers actually result in bringing them closer together rather than further apart.  A late revision to the advisory is issued.  However, the encounter ultimately results in \iac{nmac}.  \Cref{fig:cat6} shows the vertical profile of an encounter in this category.  The aircraft cross in altitude during pilot 2's response delay period at 21 seconds and \ac{nmac} occurs at 39 seconds.  

Issuing advisories can be tricky in cases where the aircraft are approximately co-altitude due to uncertainties in the aircraft's estimated positions and future intentions.  In these encounters, the problem arises as one aircraft maneuvers to cross in altitude immediately after an initial \ac{ra} is issued to the other aircraft.  Operationally, this is a rare case as the positions of the aircraft and the timing of the maneuver need to coincide.


\section{Conclusions.}

This paper introduced \ac{gbdt}, a framework that combines decision trees and \acl{gbes} for interpretable classification and categorization.  \ac{gbdt} addresses the need for interpretable models that can support heterogeneous multivariate time series data.  We validated our approach on the classic Australian Sign Language dataset from \ac{uci} and a dataset from the flight logs of an aircraft encounter simulator.  \acp{gbdt} performed well on both classification and interpretability metrics and produced highly-interpretable categorizations of the data.  The results of the study were used to inform the development of \ac{acasx}. 


\section*{Acknowledgements}

We thank Neal Suchy at the FAA, Michael Owen and Cindy McLain at MIT Lincoln Laboratory, and the \ac{acasx} team.  This work was supported by the \acs{saso} Project under NASA ARMD's \acs{aosp}.

\bibliographystyle{siam}
\bibliography{master,ritchie}

\end{document}